\documentclass{article}
\usepackage{ijcai21}

\usepackage{times}
\usepackage{soul}
\usepackage{url}
\usepackage[hidelinks]{hyperref}
\usepackage[utf8]{inputenc}
\usepackage[small]{caption}
\usepackage{graphicx}
\usepackage{amsmath}
\usepackage{amsthm}
\usepackage{booktabs}
\usepackage{algorithm}
\usepackage{algorithmic}
\usepackage{helvet}  % DO NOT CHANGE THIS
\usepackage{courier}  % DO NOT CHANGE THIS
\usepackage{natbib}  % DO NOT CHANGE THIS AND DO NOT ADD ANY OPTIONS TO IT

\usepackage{balance}
\usepackage{booktabs} % For formal tables
\usepackage{subfigure}
\usepackage{epstopdf}
\usepackage{epsfig}
\usepackage{xspace}
\usepackage{amssymb,amsmath}
\usepackage{enumitem}
\usepackage{color}
\usepackage{stfloats}
\usepackage{caption}
\usepackage{multirow}
\usepackage{float}
\usepackage[normalem]{ulem}
\useunder{\uline}{\ul}{}
\usepackage{threeparttable}
\usepackage{diagbox}
\usepackage[table]{xcolor}
\usepackage{amssymb}
\usepackage{bbding}

\newcommand{\ie}{\textit{i}.\textit{e}.}
\newcommand{\eg}{\textit{e}.\textit{g}.} 
\newcommand{\wrt}{\textit{w}.\textit{r}.\textit{t}} 

\newtheorem{Def}{Definition}

\newtheorem*{Pro*}{Problem}

\newcommand{\model}{MainTUL\xspace}
\newcommand{\modelca}{TUL-CA\xspace}
\newcommand{\modelta}{TUL-TA\xspace}

\newcommand{\modelmut}{TUL-MUT\xspace}

\title{Mutual Distillation Learning Network for Trajectory-User Linking}

\author{
Wei Chen$^1$\and
Shuzhe Li$^1$\and
Chao Huang$^2$\and
Yanwei Yu$^{1}$\footnote{Corresponding Author}\and
Yongguo Jiang$^{1}$\and
Junyu Dong$^1$\\
\affiliations
$^1$College of Computer Science and Technology, Ocean University of China\\
$^2$Department of Computer Science, The University of Hong Kong\\
\emails
\{weichen, lishuzhe\}@stu.ouc.edu.cn,
chaohuang75@gmail.com,\\
\{yuyanwei, jiangyg, dongjunyu\}@ouc.edu.cn
}

\begin{document}

\maketitle

\begin{abstract}
Trajectory-User Linking (TUL), which links trajectories to users who generate them, has been a challenging problem due to the sparsity in check-in mobility data. Existing methods ignore the utilization of historical data or rich contextual features in check-in data, resulting in poor performance for TUL task. In this paper, we propose a novel \underline{M}utu\underline{a}l d\underline{i}stillatio\underline{n} learning network to solve the \underline{TUL} problem for sparse check-in mobility data, named \textit{\model}. Specifically, \model is composed of a \textit{Recurrent Neural Network (RNN) trajectory encoder} that models sequential patterns of input trajectory and a \textit{temporal-aware Transformer trajectory encoder} that captures long-term time dependencies for the corresponding augmented historical trajectories. Then, the knowledge learned on historical trajectories is transferred between the two trajectory encoders to guide the learning of both encoders to achieve \textit{mutual distillation of information}. Experimental results on two real-world check-in mobility datasets demonstrate the superiority of \model against state-of-the-art baselines. The source code of our model is available at \url{https://github.com/Onedean/MainTUL}.
\end{abstract}

\section{Introduction}

The rapid development of Location-Based Social Network (LBSN) platforms has made it easier for humans to digitize their mobility behaviors by sharing their check-ins, opinions, and comments~\citep{lian2020geography}. These mobility behaviors can be used to understand and predict human movement patterns, facilitating intelligent business models and user experience. However, individual mobility is not always predictable due to the missing and sparsity of check-in data~\citep{lian2014analyzing}. Trajectory-user linking (TUL)~\citep{gao2017identifying} is recently proposed as a task to identify user identities based on personal mobility trajectories. It plays an important role in revealing basic human movement patterns by mining user mobility behaviors. 
% In addition, TUL could benefit a broad range of applications, such as location-sharing business services~\citep{liu2019geo}, tracking COVID-19 pandemic~\citep{hao2020understanding}, and identifying terrorists/criminals for public safety~\citep{zhou2018trajectory}. 
In addition, TUL could benefit a broad range of applications in business, transportation, epidemic prevention, and public safety, such as location-based services~\citep{liu2019geo}, tracking COVID-19 pandemic~\citep{hao2020understanding}, intelligent transportation~\cite{dai2021temporal}, and identifying terrorists/criminals for public safety~\citep{2018deepcrime}.

% TUL can essentially be seen as an extension of traditional trajectory classification tasks. Traditional trajectory measurement methods such as Longest Common Sub-Sequence (LCSS) and Dynamic Time Warping (DTW) can identify the most likely users by measuring the similarity between unknown trajectories and known trajectories. On the basis of trajectory measurement, \cite{jin2020trajectory} propose a trajectory signature approach to model the similarity of user trajectories and explore the influence of multiple factors such as spatial, temporal, spatiotemporal on moving object linking. Recently, TUL methods have been developed from trajectory similarity measures to deep trajectory representation learning based on Recurrent Neural Networks (RNNs)~\citep{gao2017identifying,zhou2018trajectory,miao2020trajectory}. 

In this work, we are interested in linking trajectories to their potential users for check-in mobility data. 
TUL can essentially be seen as an extension of traditional trajectory classification tasks. Traditional trajectory measurement methods such as Longest Common Sub-Sequence (LCSS) and Dynamic Time Warping (DTW) can identify the most likely users by measuring the similarity between unknown trajectories and known trajectories. 
Recently, a handful of studies~\citep{zhou2018trajectory,zhou2021self,miao2020trajectory} have been developed for solving the TUL problem through deep trajectory representation learning. However, the existing methods still have three key limitations. 
\textit{First,} all existing approaches still suffer from \textit{data sparsity}, and perform poorly on sparse check-in mobility datasets.  
\textit{Second,} existing methods only focus on spatial feature and/or temporal feature, and ignore the \textit{rich contextual features} in check-in data such as POI categories.  
\textit{Third,} most existing methods neglect the \textit{utilization of historical data}. Due to the inherent sparsity of check-in data, the historical data of same users implies users‘ more complex movement patterns, which may help improve the model performance. Nevertheless, how to effectively utilize the knowledge from historical data remains a significant challenge.

To address the aforementioned challenges, we propose \model, a \underline{M}utu\underline{a}l d\underline{i}stillatio\underline{n} learning network model, to solve the \underline{TUL} problem for sparse check-in trajectory data. In \model, we design two different trajectory encoders -- an RNN-based encoder to learn spatio-temporal movement patterns of input trajectory, and a temporal-aware transformer encoder to capture long-term time dependencies for the corresponding augmented trajectory. Then, the knowledge learned from the augmented trajectory data is transferred to guide the learning of RNN-based trajectory encoder. Meanwhile, input trajectory and augmented trajectory is exchanged to realize a mutual distillation learning network.
Additionally, we also design a check-in embedding layer to produce multi-semantic check-in representations integrated with POI category and time information, which are then fed into the mutual distillation learning network. 
Experimental results on two real-life human mobility datasets show that our model significantly outperforms state-of-the-art baselines (\textbf{14.95}\% Acc@1 gain and \textbf{14.11}\% Macro-F1 gain on average) in TUL task. 

Our contributions can be summarized as follows:
\begin{itemize}
    \item We propose a mutual distillation learning network model, \model, to solve TUL problem for sparse check-in trajectory data. Our \model effectively leverages history data by trajectory augmentation and knowledge distillation to improve model performance. 
    
    \item We design a temporal-aware transformer trajectory encoder in \model to capture the long-term time dependencies in the augmented trajectories. With the designed trajectory encoders, our model achieves mutual distillation of information.
    
    \item We conduct extensive experiments on two real-life check-in mobility datasets. Results show that our model significantly outperforms state-of-the-art baselines by average 14.95\% and 14.11\% improvements in terms of Acc@1 and Macro-F1.
\end{itemize}
\section{Related Work}

% A fundamental task for trajectory pattern mining is to measure the similarity or distance between trajectories. 
Trajectory similarity measures have been widely used to explore the similarity of users from their spatiotemporal trajectories. Examples include LCSS~\citep{ying2010mining}, DTW~\citep{keogh2000scaling}, Spatio-Temporal Linear Combine distance~\citep{shang2017trajectory}, and Spatiotemporal Signature~\citep{jin2019moving}. \textit{However, such measures only consider spatial or spatio-temporal proximities and cannot capture the temporal dependencies in trajectory data. }  

Recently, a variety of studies that focus on deep representation learning have been proposed for trajectory similarity computation~\citep{li2018deep,yao2019computing,yao2020linear,zhang2020trajectory,yang2021t3s}. \textit{However, these studies focus more on improving the efficiency of trajectory similarity computation.} %, rather than the measurement scheme of trajectory similarity.}  
The introduction of TUL problem~\citep{gao2017identifying,zhou2018trajectory} has further promoted the progress of deep trajectory representation learning in spatio-temporal data mining. Several methods~\citep{miao2020trajectory,zhou2021improving,gao2020adversarial} have been proposed to solve the TUL problem with deep neural networks. 
% TULER~\citep{gao2017identifying} uses an RNN-based semi-supervised learning model to exploit spatio-temporal data to capture the latent semantics of user movement patterns for TUL task. 
TULVAE~\citep{zhou2018trajectory} incorporates VAE model into TUL problem to learn hierarchical semantics of check-in trajectories in RNN. % to improve the prediction accuracy. 
DeepTUL~\citep{miao2020trajectory} proposes recurrent networks with attention mechanism to model higher-order and multi-periodic mobility patterns by learning from historical data to alleviate the data sparsity problem. 
AdattTUL~\citep{gao2020adversarial} and TGAN~\citep{zhou2021improving} introduce Generation Adversarial Network (GAN) to deal with the TUL problem. 
Recently, SML-TUL~\citep{zhou2021self} uses contrastive learning to learn the predictive representations from the user mobility itself constrained by the spatio-temporal factors. 
\textit{Nevertheless, these methods use RNNs for modeling or prediction, which cannot effectively capture long-term time dependencies, and all methods ignore the effective use of historical data.}

\cite{hinton2015distilling} first proposed the concept of Knowledge Distillation (KD) in teacher-student architecture, which seeks to provide another pathway to gain knowledge about a task by training a model with a distillation loss in addition to the task loss. %by introducing the softened output of teacher-student architecture. 
% This distillation loss is generated by the help of a teacher network, which is large in size, but achieves high accuracy on a task. The objective of distillation is to increase the accuracy of a smaller network (the student) by aiding it’s learning through this distillation loss. 
% The so-called KD is to use the complex teacher model that has been trained to guide the training of a small student model, so as to reduce the model size and computing resources while maintaining the accuracy of the original teacher model as much as possible. 
\cite{ruffy2019state} validate that appropriately tuned classical distillation in combination with a data augmentation training scheme provides orthogonal improvements. %that in turn increase model accuracy.
% In distillation task, the knowledge extraction of one model is used to improve another model, but this distillation is only unidirectional. 
Recently, \cite{zhao2021mutual} propose a trainable mutual distillation learning model, which improves end-to-end performance more effectively than traditional teacher-student framework. %, thereby improving the prediction accuracy of the model. 
\textit{In this work, we are the first attempt to use mutual distillation strategy to effectively utilize knowledge extracted from historical data to improve TUL performance.}

\section{Preliminaries}

% In this section, we first introduce some preliminary concepts and then formally define the problem of TUL. 

% Let $\mathcal{U}=\{u_1,u_2,\dots,u_i\}$ denote a set of moving users.

\begin{Def}[Check-in Record] 
A check-in record is a triple $\langle u, t, p\rangle$ that represents user $u$ visiting POI $p$ at time $t$, where $p$ denotes a uniquely identified venue in the form of $(id, category, \ell)$, $\ell$ representing the location of the POI. 
\end{Def}

% \begin{Def}[Check-in Record] 
% A check-in record is a triple $\langle u, t, p\rangle$ that represents user $u$ visiting POI $p$ at time $t$, where $p$ denotes a uniquely identified venue in the form of $(id, category, \ell)$, $\ell$ representing the geographical location of the POI (\ie, longitude and latitude). 
% \end{Def}

\begin{Def}[Trajectory] A trajectory is a sequence of check-in records $(\langle u, t_1, p_1 \rangle, \langle u, t_2, p_2 \rangle, \dots, \langle u, t_m, p_m \rangle)$ generated by user $u$ in chronological order during a certain time interval $\tau$, which is denoted by $Tr_{\tau}^{u}$.  
\end{Def} 

% \begin{Def}[Trajectory] A trajectory $Tr$ is denoted by a series of check-in points of user $u_i$ in a time interval $w_i$, \ie, $Tr_{u_i}^{w_i}=[p_1,p_2,...,p_i,...,p_l]$, where check-ins point $p$ is a record left by users accessing a certain location. Its basic elements usually include point of interest $s$ representing  spatial information and temporal information $t$. In addition, each location will have its own category attribute information $c$. Specifically, a check-in point $p=<s,t,c>$.
% \end{Def}

A trajectory is called \textit{unlinked}, if we do not know the user who generated it. The time interval $\tau$ in our work is set to 24 hours. 
We now state our problem as below: 

\begin{Pro*}[Trajectory-User Linking]
The task of TUL is to identify anonymous trajectories with the users who generate them. 
Let $\mathcal{T}=\{Tr_1,Tr_2,\dots,Tr_n\}$ represent the set of unlinked trajectories, and  $\mathcal{U}=\{u_1,u_2,\dots,u_m\}$ denote a set of users. Our goal is to find the mapping function $f(\cdot)$ satisfying the following condition:
\begin{equation}
    \label{problem_def}
    \min _{f \in \mathcal{F}} \frac{1}{n} \sum_{i=1}^{n}\left\|f(Tr_i)- y_{i}\right\|, f(Tr_i) \in \mathcal{U}, y_{i} \in \mathcal{U},
\end{equation} 
where $y_{i}$ is the true label of trajectory $Tr_i$, $\|\cdot\|$ represents difference evaluation operator (\eg, if $i=j$, $\|u_i-u_j\| = 0$, otherwise 1), and $\mathcal{F}$ is hypothesis space of TUL task. 
\end{Pro*}

% \begin{Pro*}[Trajectory-User Linking]
% The task of TUL is to identify anonymous trajectories with the users who generate them. Let $\mathcal{T}=\{Tr_1,Tr_2,\dots,Tr_m\}$ represent unlink trajectories
% set, $\mathcal{U}=\{U_1,U_2,\dots,U_n\}$ denote a set of users. Therefore, our goal is to find the mapping function $f(T)$ satisfying the following equation:

% \begin{equation}
%     \label{problem_def}
%     \min _{f \in \mathcal{F}} \frac{1}{m} \sum_{i=1}^{m}\left\|f(Tr)-Tr_{t}\right\|, f(Tr) \in \mathcal{U}, Tr_{t} \in \mathcal{U},
% \end{equation}

% where $Tr_{t}$ is the true label of trajectory $Tr$, $\|\cdot\|$ represent difference evaluation operator and $\mathcal{F}$ is hypothesis space of TUL task.

% \end{Pro*}

\section{Methodology}

\begin{figure*}
    \vspace{-2mm}
    \begin{center}
    % \hspace{-8mm}
    % \includegraphics[width=0.999\textwidth]{figure/overview2.png}
    \includegraphics[width=1.01\textwidth]{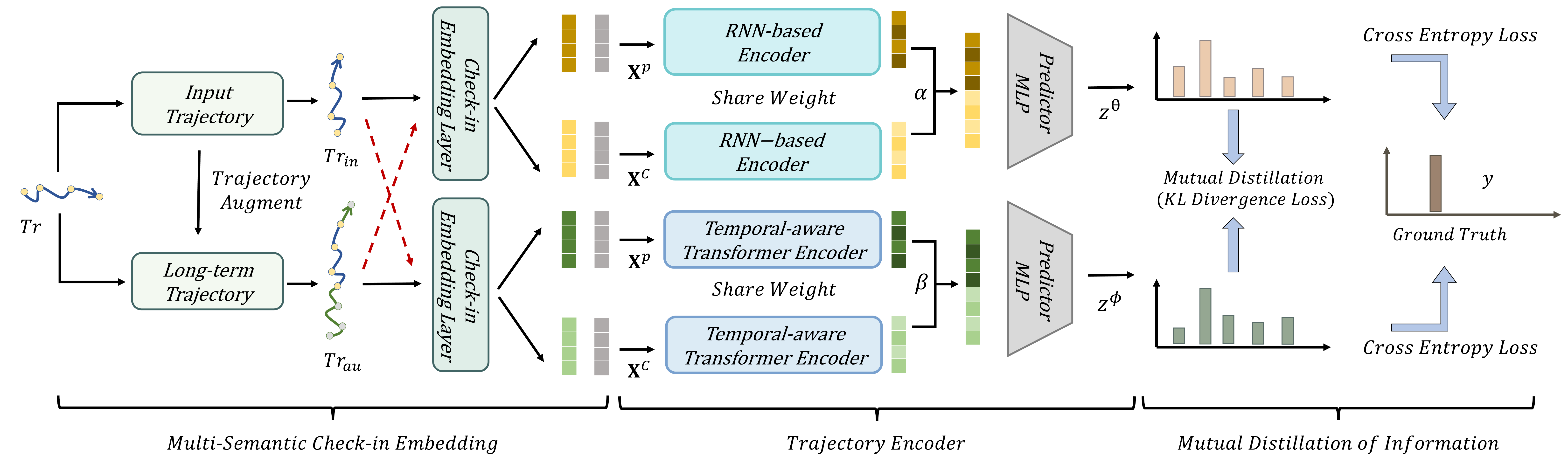}
    \vspace{-3mm}
    \caption{The architecture of the proposed \model framework.}
    \vspace{-3mm}
    \label{fig:overview}
    \end{center}
\end{figure*}

% \subsection{General Framework}

The architecture of \model is presented in Figure~\ref{fig:overview}. \model contains three major components: \textit{check-in embedding}, \textit{trajectory encoder}, and \textit{mutual distillation network}. 

% Each input trajectory is first extended to a long-term trajectory by trajectory augmentation, which is used as input together with the input trajectory. Specifically, for each input trajectory, its check-in POIs, POI categories, and time slice information are embedded to obtain \textit{multi-semantic check-in embeddings}. 
% Then, the multi-semantic embeddings of check-in sequence are fed into \textit{trajectory encoder} to extract comprehensive representative features. Particularly, \model uses a shared RNN encoder and a shared temporal-aware Transformer encoder to learn trajectory representations for the input trajectory and the augmented trajectory. 
% Finally, we leverage \textit{knowledge distillation} to exploit the dark knowledge learned by the long-term trajectory to guide the learning process of the input trajectory encoder.  Furthermore, we swap the input trajectory and the augmented trajectory branches to achieve \textit{mutual distillation of information}. 

\subsection{Multi-Semantic Check-in Embedding}

This module contains two sub-modules:  \textit{trajectory augmentation} and \textit{check-in embedding layer}.

\subsubsection{Trajectory Augmentation}

For check-in data, due to its inherent sparsity, the number of user check-ins in a sub-trajectory within a time window is very limited, while its long-term historical trajectory often implies more user movement patterns. Therefore, we explore two different trajectory augmentation strategies to generate the long-term trajectories for mutual distillation learning: 
\begin{itemize}
    \item \textit{Neighbor Augmentation:} Given an input trajectory $Tr$ in the time interval $\tau_i$, we use $k+1$ sub-trajectories in time interval $[\tau_{i-k/2}:\tau_{i+k/2}]$ to form a long-term augmented trajectory in chronological order. 
    \item  \textit{Random Augmentation:} For the input trajectory, we randomly sample $k$ sub-trajectories from all trajectories of the corresponding user, and then form a long-term trajectory together with the input trajectory according to the time information. %The purpose of this strategy is to enrich the trajectory patterns. 
\end{itemize}

%   \textit{Proximity Augmentation:} The movement trend of users is often limited by its temporal and spatial factors. Therefore, given a query trajectory $Tr$ in the time interval $w_i$, we find $k+1$ trajectories in $[w_{i-k/2}:w_{i+k/2}]$ time interval, and form a new long-term trajectory sequence according to the timing information.

%   \textit{Random Augmentation:} In this strategy, for the input trajectory, we randomly sample $k+1$ trajectories containing the current trajectory from all trajectories of the corresponding user in its training set, and then form a new long-term trajectory sequence according to the timing information. The starting point of this strategy is to enrich the trajectory sequence pattern.

\subsubsection{Check-in Embedding Layer}

The goal of multi-semantic check-in embedding is to generate dense representations for each part in a check-in (such as location, category, and time slice) from sparse one-hot representations. In this way, we not only avoid the curse of dimensionality, but also capture contextual semantic information. 
% A check-in record generally consists of a POI, its category, and a check-in timestamp. 
For each trajectory, we actually have two sequence forms, one is a sequence of POIs, and the other is a sequence of POI categories. 
% Previous studies~\citep{jin2019moving} have shown that precise timestamps would hurt the prediction performance of TUL task. 
We divide a periodic time interval (\eg, one day) into multiple time slices (\eg, one hour for one time slice) and map check-in timestamps into corresponding time slices. Finally, we separately learn the embeddings for two sequences integrating with time slices as follows: 
\begin{equation}
    \label{check-in embedding layer}
    \begin{split}
    x_i^p = tanh([\mathbf{W}_p p_i + b_p;\mathbf{W}_t t_i + b_t]),\\
    x_i^c = tanh([\mathbf{W}_c c_i + b_c;\mathbf{W}_t t_i + b_t]),
    \end{split}
\end{equation}
where $p_i$, $c_i$ and $t_i$ are one-hot encodings for the $i$-th POI, its category and time slice in the sequences, and $\mathbf{W}_p$, $\mathbf{W}_c$, $\mathbf{W}_t$, $b_p$, $b_c$ and $b_t$ are learnable parameters. %of check-in embedding layer. 

\subsection{Trajectory Encoder}

To learn higher-order transition patterns of check-in trajectory sequences, trajectory encoders need to be designed. We note that the sequence lengths of input trajectories and augmented long-term trajectories differ greatly. Overly complex trajectory encoder is not suitable for processing shorter trajectories, and overly simple trajectory encoder cannot capture the long-term time dependencies of long trajectory sequences. Therefore, we design two different trajectory encoder based on RNN model and temporal-aware self-attention network respectively.

\subsubsection{RNN-based Encoder}

RNN is an efficient architecture for processing simple variable-length sequences. Due to the sparse nature of check-in trajectory, we use a popular RNN variant, Long Short-Term Memory network (LSTM)~\citep{hochreiter1997long,2019mist}, as the encoder $f_{\theta}(\cdot)$ for processing the input trajectory. For both check-in POI and category sequences, we use a shared encoder $f_{\theta}(\cdot)$ and choose the hidden layer embedding of last time step as the representation of the two sequences. 
Then we use a learnable parameter $\alpha$ to balance the weights of the two representations and map them to user dimension via a multilayer perceptron (MLP) to obtain the final representation of the input trajectory: 
\begin{equation}
    \label{current embedding}
    z_{in}^{\theta}=MLP(\alpha \cdot f_{\theta}(\mathbf{X}_{in}^p)+(1-\alpha) \cdot f_{\theta}(\mathbf{X}_{in}^c)), 
\end{equation}
where $\mathbf{X}_{in}^p = \{x_i^p|i = 1,2,\dots,m\}$ and $\mathbf{X}_{in}^c = \{x_i^c|i = 1, 2, \\\dots, m\}$ denote the embeddings of check-in POI sequence and category sequence of input trajectory, respectively. 

\subsubsection{Temporal-aware Transformer Encoder}

The Transformer architecture~\citep{vaswani2017attention} is very expressive and flexible for both long- and short-term dependencies, which is proven to be superior to traditional sequence models in dealing with long sequences~\citep{yang2019xlnet,2020hierarchically}. However, for a long trajectory sequence, it should be noted that there is still a problem, that is, the time slice information cannot fully reflect the information changes of the trajectory in the time dimension (\eg, relative visit time difference). Therefore, inspired by~\citep{zuo2020transformer,lin2020pre}, we design a temporal-aware position encoder to replace the position encoder in the original transformer: 
\begin{equation}
    \label{temporal_encoder}
    \left[PE\left(t_{j}\right)\right]_{i}=\left\{\begin{array}{ll}
    \cos \left(w_{i} t_{j}\right), & \text { if } i \text { is odd }, \\
    \sin \left(w_{i} t_{j}\right), & \text { if } i \text { is even },
    \end{array}\right.
\end{equation}
where $w_i$ is a learnable parameter, $i$ is the order of embedding dimension ($i\leq 2d$), and $t_j$ is the visit timestamp for $j$-th check-in record. 

Therefore, for any two POIs or categories in the sequence, the relative visit time information can be captured by: 
\begin{equation}
    \label{temporal_encoder_explain}
    PE\left(t_{j}\right) PE\left(t_{j}+\Delta_{t}\right)^{\mathsf{T}}=\sum_{i=1}^{d} \cos \left(w_i\Delta_{t}\right).
\end{equation}

We use the improved temporal-aware transformer encoder as trajectory encoder $f_{\phi}(\cdot)$ for the augmented long-term trajectory. Similarly, a shared encoder $f_{\phi}(\cdot)$ is used to process the check-in POI and category sequences, and pooling the embedded tokens of last layer to obtain the latent representations of the two sequences.  Next, we also use a learnable parameter $\beta$ to balance the weights of the two representations to obtain the final representation for the augmented long-term trajectory through a MLP predictor:
\begin{equation}
    \label{long-term embedding}
    z_{au}^{\phi}=MLP(\beta \cdot f_{\phi}(\mathbf{X}_{au}^p)+(1-\beta) \cdot f_{\phi}(\mathbf{X}_{au}^c)), 
\end{equation}
where $\mathbf{X}_{au}^p$ and $\mathbf{X}_{au}^c$ denote the embeddings of check-in POI sequence and category sequence of augmented long-term trajectory, respectively.

Notice that the detailed formulations of encoders $f_{\theta}$ and $f_{\phi}$ can be found at Appendix\footnote{https://github.com/Onedean/MainTUL/tree/main/appendix}

\subsection{Mutual Distillation Network}

Different from the traditional knowledge distillation~\citep{hinton2015distilling}, in this work, we do not strictly distinguish between teacher and student networks. Both are learning from scratch rather than compressing a new network from another deeper frozen model. 

Let $Tr_{in}$ and $Tr_{au}$ represent the input trajectory and the augmented trajectory respectively. Trajectory encoder $f_\theta$ embeds $Tr_{in}$ into $z_{in}^\theta$. Similarly, $Tr_{au}$ is encoded into $z_{au}^\phi$ via encoder $f_\phi$. 
We expect encoder network $f_\theta$ can be trained similar to the true label $y$, and the knowledge of long-term trajectory with more representation ability can be transferred to $f_\theta$. During the training, the loss function $\mathcal{L}_1$ is as: 
\begin{equation}
    \begin{split}
    \label{loss 1}
    \mathcal{L}_1 = &  \mathcal{H}(y,z_{in}^\theta)+\mathcal{H}(y,z_{au}^\phi) + \\  & \lambda T^2 \operatorname{KL}(\varphi(z_{in}^\theta/T),\psi(z_{au}^\phi/T)),    
    \end{split}
\end{equation}
where $\mathcal{H(\cdot,\cdot)}$ refers to the conventional cross-entropy loss and $\operatorname{KL}(\cdot,\cdot)$ to the Kullback–Leibler divergence of softmax $\varphi$ and log-softmax $\psi$. $T$ is the temperature, intended to smooth outputs and $\lambda$ is a balancing weight.

% The above learning process may have some limitations. \textcolor{red}{Firstly, the transformer encoder network requires more data, and secondly, due to simultaneous learning, the network may tend to reinforce its error information for some input trajectories. } 
To maximize the use of training data, we propose a mutual distillation strategy, that is, exchange the input trajectory and the augmented trajectory for retraining, so that the two trajectory encoders can see more data to enhance the discriminative ability. 
In the mutual distillation strategy, encoders $f_\theta$ and $f_\phi$ are trained collaboratively and treated as peers rather than student/teacher. Specifically, for input trajectory $Tr_{in}$ and augmented trajectory $Tr_{au}$, we swap and send them to different encoders to obtain new representations $z_{in}^\phi$ and $z_{au}^\theta$, and calculate the loss function $\mathcal{L}_2$ as follows:
\begin{equation}
    \begin{split}
    \label{loss 2}
    \mathcal{L}_2 = &  \mathcal{H}(y,z_{in}^\phi)+\mathcal{H}(y,z_{au}^\theta) + \\  & \lambda T^2 \operatorname{KL}(\varphi(z_{in}^\phi/T),\psi(z_{au}^\theta/T)).    
    \end{split}
\end{equation}

This operation can also be regarded as a data augmentation strategy. Although the RNN encoder is not specially designed for long-term trajectory data, it is also beneficial to use more trajectory sequences related to the input trajectory during training. Our final loss function for optimizing mutual distillation network is: 
\begin{equation}
    \label{total_loss}
    \mathcal{L}_{total}=\mathcal{L}_1+\mathcal{L}_2.
\end{equation}

Notice that in TUL prediction stage, there is no data augmentation operation. That is, in testing, each input trajectory is encoded by trajectory encoder $f_\theta$, and then is linked with the predicted user label.

\section{Experiments}

% To verify the effectiveness of our proposed \model, we extensively compare and analyze its performance with traditional methods and deep leaning methods on TUL tasks.

%  In addition, we also verify the impact of different components on the performance of the \model and the impact of different strategies on the results. Finally, we also visually analyze the representation effects of different learning methods.

\subsection{Datasets} 

\begin{table}
    \vspace{-1mm}
    \begin{center}
    \caption{Statistics of the datasets.}
    \label{tab:dataset_table}
    \vspace{-2mm}
    \fontsize{9}{12} \selectfont
    \setlength{\tabcolsep}{0.9mm}{}	
    \begin{tabular}{c|c|c|c|c|c}
        \toprule
        \textbf{Datasets} & \#users  & \#trajectories & \#POIs & \#categories & Duration \\
        \midrule
        \multirow{2}{*}{Foursquare} & 800 & 104,413 & 39,698 & 239 & 11 months\\
        \cline{2-6}
        & 400 & 51,969 & 24,526 & 231 & 11 months\\
        \midrule
        \multirow{2}{*}{Weeplaces} & 800 & 152,583 & 24,649 & 1,373 & 7 years\\
        \cline{2-6}
        & 400 & 75,873 & 18,482 & 1,177 & 6 years\\
        \bottomrule
    \end{tabular}
    \vspace{-4mm}
    \end{center}
\end{table}

We use two real-world check-in mobility datasets~\citep{liu2014exploiting,yang2014modeling} collected from two popular location-based social network platforms, \ie,  Foursquare\footnote{http://sites.google.com/site/yangdingqi/home/foursquare-dataset} and Weeplaces\footnote{http://www.yongliu.org/datasets.html}. 
% For each user, we connect all check-in records in chronological order to form a trajectory, which will be further divided into sub-trajectories according to the given time interval (\ie, one day). 
%Most users have very sparse check-ins in the original dataset, which makes TUL impossible for those users who have only a few check-ins over a long period of time (\eg, a month). %The amount of data in previous studies is very small, which often causes deep learning-based models to be easily saturated. 
For Foursquare and Weeplaces, we choose top 800 and 400 users with the most check-ins for evaluating model performance respectively. In experiments, we use the first 80\% of sub-trajectories of each user for training and the remaining 20\% for testing, and select 20\% training data as the validation set to cooperate with the early stop mechanism to find the best parameters and avoid overfitting. 

The statistics of two datasets are summarized in Table~\ref{tab:dataset_table}.

\subsection{Baselines}

% We compare our \model against the following three categories of baseline methods.

We consider the following baselines for comparison.
\begin{itemize}

 \item \textbf{Classical methods:} (1) \textbf{LCSS} -- a common and effective trajectory similarity measure method~\citep{ying2010mining}. (2) Signature Representation (\textbf{SR}) --  a state-of-the-art trajectory similarity measure for moving object linking~\citep{jin2019moving,jin2020trajectory}.

 \item \textbf{Machine learning methods:} (3) Linear Discriminant Analysis (\textbf{LDA}) --  a classic spatial data classification method~\citep{shahdoosti2017spectral}. (4) Decision Tree (\textbf{DT}) -- an effective classification method for trajectory data~\citep{jiang2018survey}. 

 \item \textbf{Deep neural network models:} (5) \textbf{TULER} -- an RNN model for TUL task~\citep{gao2017identifying}, including three variants: RNN with Gated Recurrent Unit (\textbf{TULER-G}), LSTM (\textbf{TULER-L}) and bidirectional LSTM (\textbf{Bi-TULER}). (6) \textbf{TULVAE} -- It utilizes VAE to learn the hierarchical semantics of trajectory in RNN~\citep{zhou2018trajectory}. (7) \textbf{DeepTUL} -- a recurrent network with attention mechanism for TUL task~\citep{miao2020trajectory}.
\end{itemize}

\subsection{Evaluation Metrics and Parameter Settings}

We use the Acc@$k$, Macro-Precision, Macro-Recall and Macro-F1 to evaluate the model performance. Specifically, Acc@$k$ is used to evaluate the accuracy of TUL prediction. 
% $Macro\text{-}F1$ takes into account both the precision $Macro\text{-}P$ and recall $Macro\text{-}R$ of the model.
% Macro-F1 is the harmonic mean of \textit{precision} and \textit{recall} across all classes, which can be regarded as an overall performance indicator. 

% \subsubsection{Implementation Details}

% All experiments are conducted on a machine with Intel Xeon Gold 6126 @2.60GHz
% 12 cores CPU and 8 $\times$ NVIDIA Tesla V100-SXM2 (16GB Memory) GPU. xxx

For baselines, we use the parameter settings recommended in their papers and fine-tune them to be optimal. For \model, we set check-in embedding dimension $d$ to 512, $\lambda$ to 10, use early stopping mechanism, and set patience to 3 to avoid over fitting. The learning rate is initially set to 0.001 and decays by 10\% every 5 epochs. We repeat 10 runs for each experiment and report the average for all methods. More experimental settings can be found in the appendix. 

\subsection{Overall Performance}

\begin{table*}[t]
\vspace{-2mm}
\centering
\footnotesize
\caption{Performance comparison with deep neural network models. Macro-P/R: Macro-Precision/Recall}
\vspace{-2mm}
\label{tab:deep_perform_compared}
\fontsize{9}{12} \selectfont
\setlength{\tabcolsep}{1.5mm}{}	
\begin{tabular}{c|c|c|c|c|c|c|c|c|c|c|c}
\toprule

\multirow{2}{*}{Dataset} & \multirow{2}{*}{Methods} & Acc@1 & Acc@5 & Macro-P & Macro-R & Macro-F1 & Acc@1 & Acc@5 & Macro-P & Macro-R & Macro-F1\\

\cline{3-12}

& & \multicolumn{5}{c|}{$|\mathcal{U}|=400$} & \multicolumn{5}{c}{$|\mathcal{U}|=800$}\\

\midrule

\multirow{6}{*}{Foursquare} &TULER-L &52.88\%  &64.08\%  &58.96\%  &50.69\%  &52.75\%  &47.36\%  &59.11\%  &52.52\%  &45.28\%  &46.73\%\\

\cline{2-12}

&TULER-G &52.41\%  &63.41\%  &58.39\%  &50.17\%  &52.09\%  &47.67\%  &58.94\%  &52.39\%  &45.53\%  &46.55\%\\

\cline{2-12}

&Bi-TULER &51.59\%  &63.02\%  &58.46\%  &49.45\%  &51.62\%  &\underline{49.03}\%  &\underline{60.48}\%  &54.92\%  &\underline{47.10}\%  &\underline{48.51}\%\\

\cline{2-12}

&TULVAE &52.61\%  &63.33\%  &58.75\%  &50.51\%  &52.66\%  &47.71\%  &59.55\%  &53.20\%  &45.65\%  &47.09\%\\

\cline{2-12}

&DeepTUL &\underline{53.99}\%  &\underline{65.22}\%  &\underline{61.78}\%  &\underline{52.00}\%  &\underline{54.65}\%  &45.41\%  &58.20\%  &\underline{57.75}\%  &43.33\%  &45.43\%\\

\cline{2-12}

& \textbf{\model} &\textbf{60.58}\%  &\textbf{70.22}\%  &\textbf{62.91}\%  &\textbf{58.49}\%  &\textbf{59.12}\%  &\textbf{56.92}\%  &\textbf{67.85}\%  &\textbf{59.20}\%  &\textbf{55.02}\%  &\textbf{55.37}\%\\

\midrule

\multirow{6}{*}{Weeplaces} &TULER-L &38.44\%  &49.60\%  &41.54\%  &37.50\%  &37.33\%  &36.12\%  &47.56\%  &40.62\%  &35.27\%  &35.78\%\\

\cline{2-12}

&TULER-G &38.84\%  &50.17\%  &42.56\%  &37.76\%  &38.09\%  &\underline{36.35}\%  &47.78\%  &40.63\%  &\underline{35.40}\%  &\underline{35.81}\%\\

\cline{2-12}

&Bi-TULER &38.15\%  &50.21\%  &43.23\%  &37.19\%  &37.77\%  &36.23\%  &\underline{47.93}\%  &\underline{40.89}\%  &35.29\%  &35.60\%\\

\cline{2-12}

&TULVAE &\underline{38.98}\%  &\underline{50.60}\%  &42.59\%  &\underline{38.11}\%  &\underline{38.36}\%  &35.23\%  &46.84\%  &38.87\%  &34.32\%  &34.51\%\\

\cline{2-12}

&DeepTUL &33.48\%  &46.34\%  &\underline{46.50}\%  &33.08\%  &34.37\%  &26.65\%  &37.73\%  &37.22\%  &25.66\%  &26.24\%\\

\cline{2-12}

&\textbf{\model} &\textbf{45.31}\%  &\textbf{58.28}\%  &\textbf{49.81}\%  &\textbf{44.24}\%  &\textbf{45.22}\%  &\textbf{41.90}\%  &\textbf{55.51}\%  &\textbf{46.63}\%  &\textbf{40.58}\%  &\textbf{41.62}\%\\

\bottomrule

\end{tabular}
\vspace{-2mm}
\end{table*}

\begin{figure}[b]
\vspace{-3mm}
\centering
% \hspace{-5mm}
\includegraphics[width=0.9\columnwidth]{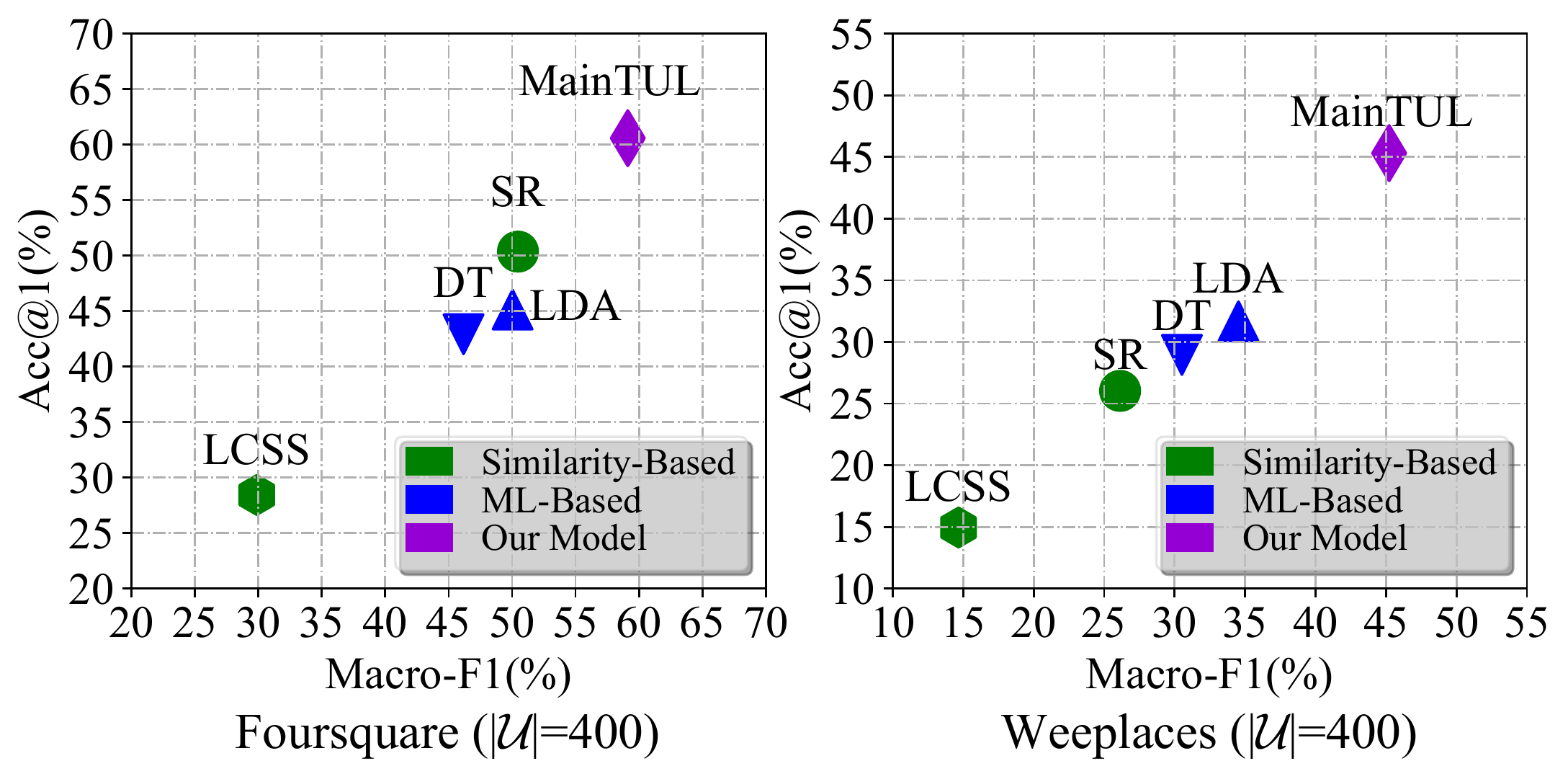}
\vspace{-3mm}
\caption{Performance comparison with classical models.}
\label{fig:compared_traditional}
\vspace{-3mm}
\end{figure}

% \subsection{Results compared with deep learning methods}

We report the overall performance with deep neural network models in Table ~\ref{tab:deep_perform_compared}, where the best is shown in bold and the second best is shown as underlined. The comparison with classical learning methods are shown in Figure~\ref{fig:compared_traditional}. 
 
As shown in Table ~\ref{tab:deep_perform_compared} and Figure~\ref{fig:compared_traditional}, \model significantly outperforms all baselines in terms of all evaluation metrics on two real-world check-in datasets. Specifically, \model achieves average 14.95\% Acc@1 and 14.11\% Macro-F1 improvement in comparison to the bested performed baseline on two datasets. The main reason is that our designed mutual distillation model based on two different trajectory encoders captures the spatio-temporal movement patterns of users' check-in trajectories more effectively than RNN-based models (\eg, TULER and DeepTUL). 
In addition, contextual features such as POI category and temporal information are integrated in \model to further improve the performance.

We also observe that model performance on data with more users is worse than that on data with fewer users. This is intuitive because the more users the more difficult the classification becomes. However, the performance of our \model is only reduced by 3.68\% in Macro-F1 from $|\mathcal{U}|=400$ to $|\mathcal{U}|=800$, while state-of-the-art model (\ie, DeepTUL) is reduced by 8.13\%-9.22\%. 
For DeepTUL, considering the historical data of all users does have a certain improvement on the data with fewer users, but when the number of users is large, a large amount of history data will also bring more noise, resulting in a sharp drop in performance. 
However, our model only uses the historical data of same user for trajectory enhancement and knowledge distillation during training, and does not require historical data for testing, and thus still performs better on the data with more users.

Notice that SML-TUL~\citep{zhou2021self} and TGAN \citep{zhou2021improving} are not compared in our experiments due to no publicly available source codes. However, our \model significantly outperforms SML-TUL and TGAN in terms of Acc@$k$ and Macro-F1 on Foursquare according to the results reported in~\citep{zhou2021self}, even if \model links more users (\eg, \underline{\model} vs. SML-TUL vs. TGAN: \underline{60.58\%} vs. 57.23\% vs. 53.00\% in Acc@1, and \underline{59.12\%} vs 52.66\% vs. 47.76\% in Macro-F1 on Foursquare).

\subsection{Ablation Study}

\subsubsection{Component Ablation}

% We next conduct the ablation study to verify the effectiveness of each component in \model. We compare our model with the following four carefully designed variations.  Despite the changed part(s), all variations have the same framework structure and parameter settings. 

% We next conduct the ablation study to verify the effectiveness of each component in \model. 
In our ablation study, we compare our model with the following three carefully designed variations: (1) \textbf{\modelca} -- This variation removes the category and time information in check-in embedding layer. (2) \textbf{\modelta} -- It uses position encoding in~\citep{vaswani2017attention} to replace our proposed temporal-aware position encoding. (3) \textbf{\modelmut} -- It removes loss function $\mathcal{L}_2$ to prove the importance of extracting knowledge from each other.

% \begin{itemize}
%     \item \textbf{\modelca} -- This variation removes the category and time information in check-in embedding layer to demonstrate the necessity of multi-semantic contextual features.
    
%     % \item \textbf{\modelau} -- It removes the trajectory augmentation module and only feeds the input trajectory into two encoders, which verifies the importance of long-term augmented trajectories. 
    
%     \item \textbf{\modelta} -- It uses position encoding in~\citep{vaswani2017attention} to replace our proposed temporal-aware position encoding.
    
%     \item \textbf{\modeldis} -- In this variation, we remove the mutual distillation operation to prove the importance of extracting knowledge from each other. 
% \end{itemize}

\begin{figure}
\vspace{-2mm}
\centering
% \hspace{-6mm}
\includegraphics[width=0.9\columnwidth]{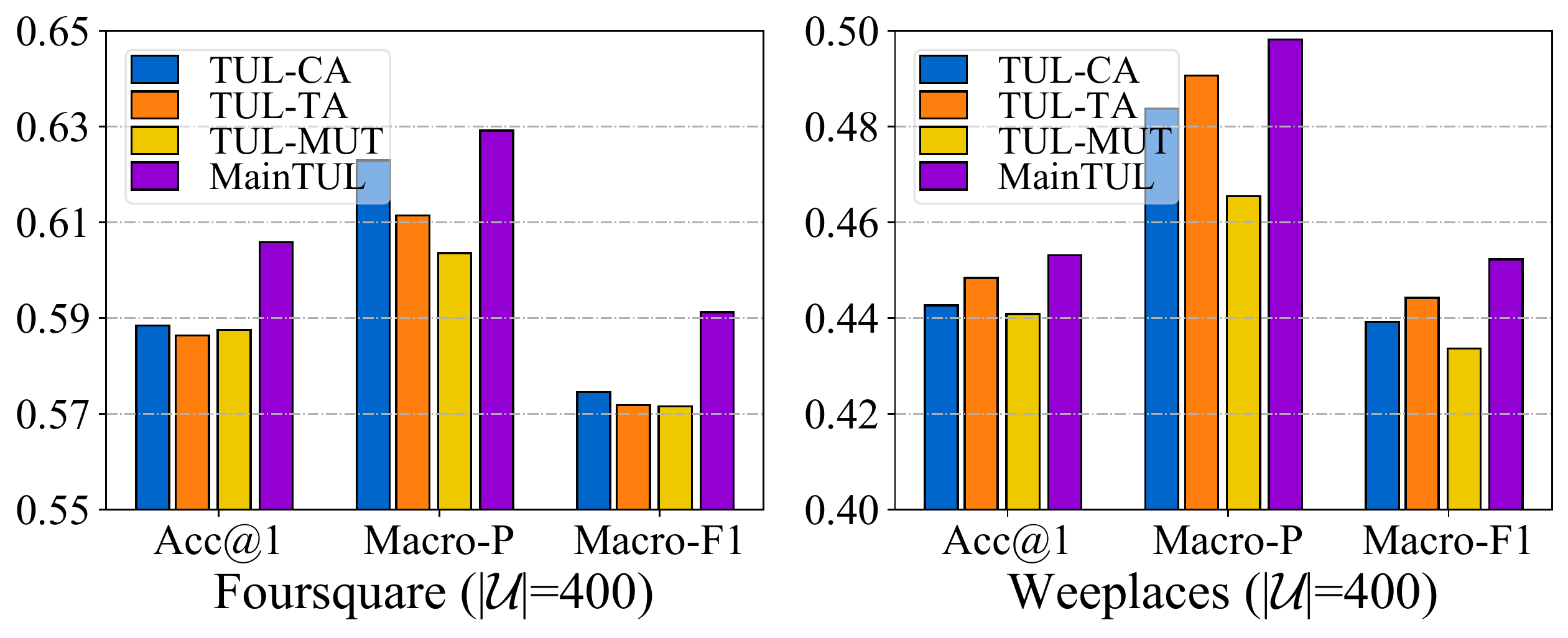}
\vspace{-3mm}
\caption{Experimental results of component ablation.}
\label{fig:ablation}
\vspace{-1mm}
\end{figure}

The results of ablation study are shown in Figure~\ref{fig:ablation}.  
As we can see, the key components all contribute to performance improvement of our model. 
% The comparison between \modeldis and \model highlights the effectiveness of the mutual distillation strategy.
We can observe that \modelmut performs the worst in most cases, indicating that the mutual distillation strategy has the greatest impact on the performance improvement of our model. 
The comparisons between \modelca, \modelta and \model reflects the importance of the contextual features and temporal-aware position encoding, respectively. The results in Figure~\ref{fig:ablation} demonstrate that temporal-aware position encoding has a greater impact on model performance on Foursquare dataset, while contextual features have a greater effect on Weeplaces dataset. 

% As we can see, the replaced components all contribute to the model, the mutual distillation strategy have a great influence, because the learning ability of a single model is limited, and it is limited to the data. The category and time of trajectory have a certain impact, because the POIs contain the main human movement patterns. In addition, the timing aware encoder is also helpful for long-term trajectory timing modeling. 

\begin{table}[h]
    \vspace{-1mm}
    \caption{Loss function explorations. Mac-F1: Macro-F1.}
    \label{tab:LossStudy}
    \vspace{-3mm}
    \newcommand{\tabincell}[2]{\begin{tabular}{@{}#1@{}}#2\end{tabular}}
    \centering
    \setlength{\tabcolsep}{1.2mm}\fontsize{9}{12}\selectfont
    \begin{tabular}{lcc}
    \toprule
    Loss function & \tabincell{c}{Foursqaure\\Acc@1/Mac-F1} & \tabincell{c}{Weeplaces\\Acc@1/Mac-F1}\\\midrule \rowcolor{gray!30}
    original & 60.58\%/59.12\% & 45.31\%/45.22\%\\
    w/o all $\lambda T^2 \operatorname{KL}(\cdot)$ & 55.36\%/53.57\% & 40.01\%/39.24\%\\
    
    w/o $\mathcal{H}(y,z_{in}^\theta)$ and $\mathcal{H}(y,z_{in}^\phi)$ & 60.06\%/58.64\% & 43.49\%/43.58\%\\
    \bottomrule
    \end{tabular}
    \vspace{-2mm}
\end{table}

\subsubsection{Loss Function Ablation}

We also evaluate the effectiveness of each term in our loss function (Eq.~\eqref{total_loss}). 
Table~\ref{tab:LossStudy} depicts the experimental results of different loss functions on two datasets with 400 users. 

First, we can see that removing term $\lambda T^2 \operatorname{KL}(\cdot)$ leads to the  decrease in performance, which demonstrates the importance of dark knowledge of long-term augmented trajectory. Second, we notice that removing term $\mathcal{H}(y,z_{in}^\theta)$ and $\mathcal{H}(y,z_{in}^\phi)$, i.e., without considering the input trajectory labels, the model performance does not drop significantly. This indicates that our model can still achieve better results only through the knowledge distillation of augmented trajectories. 

% First, we find that removing the $\lambda T^2 \operatorname{KL}(\cdot)$ term, the model performance will decrease sharply, which shows the importance of dark knowledge of long-term trajectory. 
% In addition, we noticed that the removal of $\mathcal{H}(y,z_{in}^\theta)$ and $\mathcal{H}(y,z_{in}^\phi)$ term means that the model does not know the true label of short-term trajectory at all, but due to the design of mutual distillation, it can still learn a good model only through the knowledge of distillation.

\vspace{-1mm}
\subsection{Parameter Study}

\begin{figure}
\centering
\vspace{-2mm}
\includegraphics[width=0.9\columnwidth]{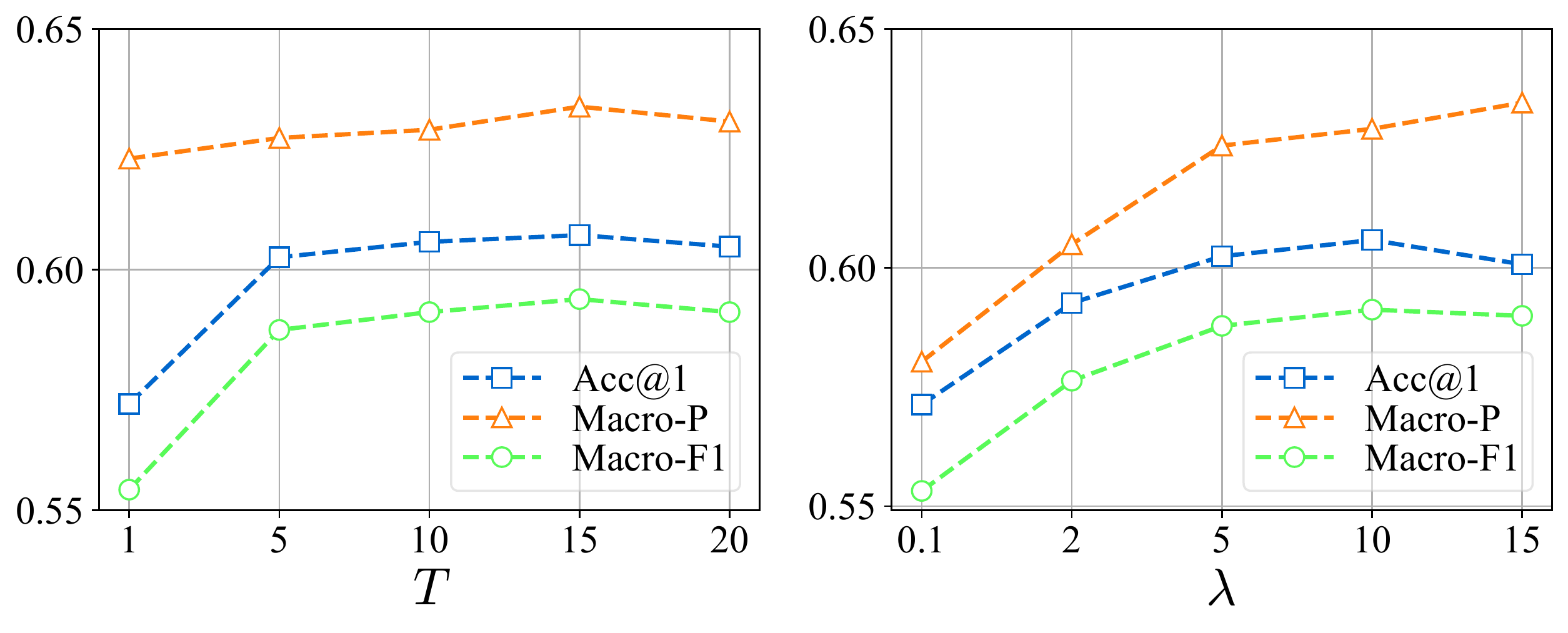}\\
\vspace{-3mm}
\caption{Parameter sensitivity \wrt. $T$ and $\lambda$ on Foursquare.}
\label{fig-parm}
\vspace{-3mm}
\end{figure}

We also evaluate the sensitivity of our \model with respect to different settings of temperature $T$ and hyperparameter $\lambda$ in our loss function. The results on Foursquare are shown in Figure~\ref{fig-parm}. 
As we can see, the performance first increases and then decreases, as $T$ increases. This is intuitive because lower temperature the distribution more sharp, and higher temperature makes the distribution unable to extract effective information. 
The same observation is also presented on hyperparameter $\lambda$. 
In addition, it is clear that the performance increases rapidly with $\lambda$ increasing from 0.1. This suggests our proposed knowledge distillation module contributes a lot to the overall performance. 

% The results show that the $T$ increase first and then decrease, which is understandable. About low $T$ makes the distribution more sharp, and about high $T$ makes the distribution unable to extract effective information. The weight $\lambda$ of distillation loss is a relatively robust parameter selection in our model, with only marginal difference in performance. Even tests conducted with an $\lambda$ of 0.1 also achieved significantly higher performance than none (see Table~\ref{tab:LossStudy}).

\subsection{Strategy Study}

\begin{table}[h]
    \vspace{-2mm}
    \caption{Effect of trajectory encoder selection. R-based: RNN-based, T-based: Transformer-based.}
    \label{tab:EncoderStudy}
    \vspace{-3mm}
    \centering
    \setlength{\tabcolsep}{1.1mm}\fontsize{9}{12}\selectfont
    \begin{tabular}{cc|cc|cc|cc}
    \toprule
    \multicolumn{4}{c|}{Train} & \multicolumn{2}{c|}{Test} & \multirow{3}{*}{Acc@1} & \multirow{3}{*}{Mac-F1}\\\cline{1-6}
    \multicolumn{2}{c|}{$f_\theta$} & \multicolumn{2}{c|}{$f_\phi$} & \multirow{2}{*}{$f_\theta$} & \multirow{2}{*}{$f_\phi$} &  & \\\cline{1-4}
    R-based & T-based & R-based & T-based &  &  & & \\\midrule
    \checkmark &  & \checkmark &  & \checkmark & & 56.47\% & 55.03\%\\ 
    
    \checkmark &  & \checkmark &  &  & \checkmark & 56.55\% & 55.23\%\\
    
    \rowcolor{gray!40} \checkmark &  &  & \checkmark & \checkmark & & 60.58\% & 59.12\%\\
    
    \rowcolor{gray!15} \checkmark &  &  & \checkmark & & \checkmark & 57.87\% & 55.83\%\\
    
    & \checkmark & \checkmark &  & \checkmark & & 57.60\% & 55.75\%\\
    
    & \checkmark & \checkmark &  & & \checkmark & 59.39\% & 58.22\%\\
    
    & \checkmark & & \checkmark & \checkmark & & 53.80\% & 53.56\%\\
    
    & \checkmark & & \checkmark & & \checkmark & 53.80\% & 53.94\%\\
    \bottomrule
    \end{tabular}
    \vspace{-3mm}
\end{table}

\subsubsection{Effect of Trajectory Encoder Selection}

We also evaluated the combination selection of different types of trajectory encoders. The results on Foursquare are shown in Table~\ref{tab:EncoderStudy}. 
We can conclude that when using the same type of trajectory encoders at the same time, the model performance is poor, which validates the rationality of our design of two different types of trajectory encoders.
In the case of using two different encoders (\ie, $f_{\theta}$ uses RNN encoder and $f_{\phi}$ adopts transformer encoder), using encoder $f_{\theta}$ for testing can achieve the best performance, which also shows that the simple encoder is more suitable for capturing the movement patterns of sparse sub-trajectories.

\subsubsection{Effect of Augmentation Strategy}

Finally, we evaluate the proposed two trajectory augmentation strategies. 
Based on the results on Foursquare in Table~\ref{tab:AugmentationStudy}, we have two observations: (1) Both data augmentation strategies help improve model performance. 
(2) The random augmentation is better than the neighbor augmentation. The reason is that the random strategy may obtain the potential movement patterns of users by combining users' past travels randomly, resulting in better results. 
% (3) The effect of increasing neighbors $k$ on model performance gradually becomes better, because longer augmented trajectories would bring richer knowledge of movement patterns. 

\begin{table}[h]
    \vspace{-2mm}
    \caption{Effect of augmentation strategy in Macro-F1.}
    \label{tab:AugmentationStudy}
    \vspace{-2mm}
    \centering
    \setlength{\tabcolsep}{1.2mm}
    \begin{tabular}{c|c|c|c|c|c}
    \toprule
    $k$ & 0 & 2 & 4 & 8  & 16\\
    \midrule
    Neighbor & 51.09\% & 53.78\% & 54.01\% & 53.49\% & 53.51\%\\
    \rowcolor{gray!30} Random & 51.09\% & 54.25\% & 59.14\% & 59.12\% & 59.00\%\\
    \bottomrule
    \end{tabular}\vspace{0cm}
    \vspace{-3mm}
\end{table}

% \begin{table}[h]
%     \caption{Effect of Augmentation Strategy. Metric: Macro-F1.}
%     \label{tab:AugmentationStudy}
%     \vspace{-4mm}
%     \centering
%     \setlength{\tabcolsep}{0.9mm}\fontsize{9}{12}\selectfont
%     \begin{tabular}{c|c|c|c|c|c|c}
%     \hline
%     \diagbox [width=5em,trim=l]{Strategy}{$k$} & 0 & 2 & 4 & 8  & 16 & 32\\
%     \hline
%     Neighbor & 51.09\% & 53.78\% & 54.01\% & 53.49\% & 53.51\% & 56.31\%\\
%     \hline
%     Random & 51.09\% & 54.25\% & 59.14\% & 59.12\% & 59.00\% & 53.49\%\\
    
%     \hline
%     \end{tabular}\vspace{0cm}
%     \vspace{-5mm}
% \end{table}

% \subsection{Case Study}

\section{Conclusion}

In this paper, we propose a novel mutual distillation learning network (\model) to solve TUL problem for sparse check-in mobility data. 
\model effectively learns user movement patterns for input trajectory by the designed mutual distillation network consisting of two different trajectory encoders with multi-semantic check-in embeddings. 
% Additionally, \model also considers the POI category and time information in check-in trajectory sequences to generate multi-semantic check-in embedding vectors, which are then fed into the mutual distillation network 
Experiments on two real-world check-in mobility datasets demonstrate that \model significantly outperforms state-of-the-art baselines in terms of all evaluation metrics for TUL prediction.

\section*{Acknowledgments}
This work is partially supported by the National Natural Science Foundation of China under grant Nos. 62176243, 61773331 and 41927805, the National Program on Key Research Project  under grant No. 2019YFC1509100, and the National Key Research and Development Program of China under grant No. 2018AAA0100602.

\clearpage
% \newpage
\bibliographystyle{named}
\bibliography{ijcai21}

\end{document}